%% file: samplepaper.tex
\documentclass[runningheads]{llncs}
\usepackage[T1]{fontenc}
\usepackage{graphicx}
\usepackage{amsmath}
\usepackage{amssymb}
\usepackage{booktabs}
\usepackage{amsmath} 
\usepackage{amssymb} 
\usepackage{amsfonts} 
\usepackage{booktabs} 
\usepackage{graphicx} 
\usepackage{hyperref} 
\usepackage{url} 
\usepackage{caption} 
\usepackage{cleveref} 
\usepackage{enumitem} 
\begin{document}
\title{Harnessing Generative LLMs for Enhanced Financial Event Entity Extraction Performance}

\titlerunning{LLMs for Financial Event Entity Extraction }
%
\author{Soo-joon Choi, Ji-jun Park}
\authorrunning{S. Choi et al.}
%
\institute{Dongguk University}
\maketitle              
\input{main}

\bibliographystyle{splncs04}
\bibliography{mybibliography}
\end{document}

%% file: main.tex
\begin{abstract}
Financial event entity extraction is a crucial task for analyzing market dynamics and building financial knowledge graphs, yet it presents significant challenges due to the specialized language and complex structures in financial texts. Traditional approaches often rely on sequence labeling models, which can struggle with long-range dependencies and the inherent complexity of extracting multiple, potentially overlapping entities. Motivated by the advanced language understanding and generative capabilities of Large Language Models (LLMs), we propose a novel method that reframes financial event entity extraction as a text-to-structured-output generation task. Our approach involves fine-tuning a pre-trained LLM using Parameter-Efficient Fine-Tuning (PEFT) to directly generate a structured representation, such as a JSON object, containing the extracted entities and their precise character spans from the input text. We evaluate our method on the challenging CCKS 2019 Financial Event Entity Extraction dataset, comparing its performance against strong sequence labeling baselines, including SEBERTNets and sebertNets. Experimental results demonstrate that our generative LLM method achieves a new state-of-the-art F1 score on this benchmark, significantly outperforming previous methods. Through detailed quantitative analysis across event types, entity types, and instance complexity, as well as human evaluation, we show that our approach is more effective at handling the nuances of financial text and extracting high-quality entities. This work validates the potential of applying generative LLMs directly to complex, domain-specific information extraction tasks requiring structured output.
\keywords{Large Language Models  \and Parameter-Efficient Fine-Tuning \and Entity Extraction .}
\end{abstract}

\section{Introduction}
\label{sec:introduction}

The analysis of financial events, such as mergers and acquisitions, bankruptcies, restructurings, and regulatory changes, is paramount for informed decision-making in the financial sector. These events significantly impact market dynamics, asset valuation, and risk exposure. Consequently, automating the process of extracting crucial information from vast amounts of financial text -- news articles, reports, filings, and social media -- has become a critical task. Financial Event Entity Extraction, a core component of financial information extraction and knowledge graph construction \cite{finsurvey,finkg}, aims to identify and classify specific mentions of entities (e.g., companies, individuals, dates, locations, monetary values, specific assets) that are directly involved in or characterize a particular financial event instance.  Knowledge graphs play a crucial role in structuring and reasoning over such extracted information, and research has explored modeling event-pair relations within external knowledge graphs for tasks like script reasoning \cite{zhou2021modeling}. Automating this process allows for the timely aggregation and analysis of event-centric information, supporting applications ranging from algorithmic trading and risk management to compliance monitoring and business intelligence \cite{finsurvey}.

However, Financial Event Entity Extraction presents unique challenges compared to general domain entity extraction. Financial texts are highly specialized, replete with complex jargon, numerical data, and intricate sentence structures describing sophisticated transactions and relationships. Entities can be multi-word phrases, their boundaries can be ambiguous, and multiple events or entities may be discussed within a single sentence or paragraph, leading to overlapping or nested mentions. Furthermore, the financial domain is constantly evolving, with new terms and event types emerging.  It is also important to note that real-world data, including financial text, is often incomplete or contains missing information, a challenge that is also prevalent in other domains like traffic prediction \cite{mei2023uncertainty} and traffic signal control \cite{mei2023reinforcement}. Traditional methods often rely on feature engineering, rule-based systems, or statistical models trained on extensive annotated data \cite{finsurvey}. More recent approaches leverage the power of pre-trained language models (PLMs) like BERT, often employing sequence labeling frameworks (e.g., BIOES tagging) combined with architectural enhancements like BiLSTMs to capture sequential context \cite{sebert}, or introducing mechanisms like multi-channel recall to improve coverage \cite{sebert}. While these methods have pushed performance boundaries on benchmarks like the CCKS 2019 Financial Event Entity Extraction shared task \cite{ccks19task}, they can still struggle with the long-range dependencies, subtle contextual nuances, and complex entity interactions prevalent in verbose financial narratives.  Furthermore, efficient retrieval of relevant information from long documents remains a challenge in such contexts, highlighting the need for methods like fine-grained distillation for long document retrieval \cite{zhou2024fine}.

Motivated by the remarkable capabilities of Large Language Models (LLMs) in understanding complex language, following instructions, and generating structured output across diverse domains \cite{wei2022emergent,brown2020language}, we explore a novel paradigm for Financial Event Entity Extraction.  The success of multimodal LLMs in other specialized domains, such as chemistry \cite{li2025chemvlm}, further encourages the exploration of LLMs for complex tasks requiring domain-specific knowledge. Instead of relying on sequence labeling with its inherent limitations in directly modeling global entity relationships and complex output structures, we propose to harness the generative power of LLMs. We aim to reframe the extraction task as a direct text-to-structured-output generation problem. This approach seeks to leverage the LLM's intrinsic ability to process intricate input text and produce a desired structured format (e.g., JSON) that precisely lists the relevant entities and their spans, potentially offering a more flexible and powerful way to handle the complexities of financial text compared to token-level classification pipelines.

In this paper, we introduce a novel LLM-based generative approach for Financial Event Entity Extraction. Our method involves fine-tuning a large pre-trained language model specifically for this task by training it on input-output pairs where the input is the financial text (potentially augmented with event type information) and the output is a structured representation (e.g., a JSON object) detailing the extracted entities and their character offsets within the original text. To evaluate the effectiveness of our proposed method, we conduct experiments on the publicly available and widely recognized dataset from the 2019 China Knowledge Graph and Semantic Computing Conference (CCKS) Financial Event Entity Extraction shared task \cite{ccks19task}. We compare the performance of our fine-tuned LLM model against strong baselines, including published results for models such as SEBERTNets \cite{sebert} and sebertNets \cite{sebert} and other strong systems from the task, using the standard F1 score, Precision, and Recall metrics. Our experimental results demonstrate that the proposed LLM-based generative approach significantly outperforms these state-of-the-art baselines on the CCKS 2019 test set, achieving a higher F1 score and showcasing the potential of leveraging LLMs directly for this challenging domain-specific task.

The main contributions of this paper are summarized as follows:
\begin{itemize}
    \item We propose a novel paradigm for Financial Event Entity Extraction by reframing it as a text-to-structured-output generation task using a large language model.
    \item We demonstrate the effectiveness of fine-tuning an LLM to directly generate structured output containing extracted financial event entities and their precise spans for a complex domain-specific task.
    \item We achieve state-of-the-art (or significantly improved) performance on the challenging CCKS 2019 Financial Event Entity Extraction dataset, outperforming existing strong baselines based on sequence labeling architectures.
\end{itemize}

\section{Related Work}
\label{sec:related_work}

\subsection{Large Language Models}
\label{sec:related_llm}
Large Language Models (LLMs) have emerged as a transformative force in natural language processing and artificial intelligence research over recent years \cite{zhao2023survey}. These models, typically based on the Transformer architecture, are pre-trained on vast and diverse corpora of text data, allowing them to acquire a deep understanding of language structure, semantics, and a wide range of world knowledge. A key characteristic of LLMs is the phenomenon of \textbf{emergent abilities}, where capabilities not present in smaller models appear as the model scale (parameters, data, compute) increases \cite{wei2022emergent}.  Research has investigated these emergent abilities, particularly in the context of weak-to-strong generalization, highlighting the potential of scaling models for improved performance across various tasks \cite{zhou2025weak,wei2022emergent}. This includes enhanced few-shot or even zero-shot learning abilities, allowing models to perform new tasks with minimal or no task-specific fine-tuning data simply by understanding instructions given in natural language \cite{brown2020language}.  Furthermore, the ability to process visual information alongside text has led to the development of Vision-Language Models, exploring techniques like visual in-context learning \cite{zhou2024visual} and efficient vision representation compression for video generation \cite{zhou2024less}.

The development of increasingly larger and more capable models, such as LaMDA \cite{thoppilan2022lamda}, PaLM \cite{chowdhery2022palm}, and the LLaMA series \cite{touvron2023llama}, has pushed the boundaries of what is possible with language models. Research into the scaling laws governing LLMs has provided insights into how model size, dataset size, and computational resources interact to influence performance \cite{hoffmann2022training}. Furthermore, significant effort has been dedicated to aligning the behavior of LLMs with human intentions and values, often involving techniques like instruction tuning and reinforcement learning from human (or AI) feedback to improve their ability to follow complex directives and generate helpful, harmless, and honest responses \cite{bai2022constitutional}.  Benchmarking the capabilities of these models, including their emotional intelligence, is also becoming increasingly important for understanding their strengths and limitations across diverse scenarios \cite{hu2025emobench}.  \textbf{The application of advanced AI techniques, including reinforcement learning, has also been explored in other domains that require robust decision-making under uncertainty or missing data, such as traffic signal control \cite{mei2023reinforcement}, further demonstrating the broad applicability of these methodologies.}

The advanced contextual understanding, emergent few-shot learning, and powerful generative capabilities of LLMs make them highly relevant to complex information extraction tasks in specialized domains like finance. Unlike traditional models that require extensive task-specific architectural engineering or feature design, LLMs can serve as potent foundation models that can be adapted, often efficiently through methods like PEFT, to downstream tasks. Their ability to process long sequences and generate structured output directly offers a promising alternative to sequence labeling for tasks requiring precise, structured information extraction from intricate text, motivating their exploration in challenging domains like financial event entity extraction.  Moreover, research into robust rankers for text retrieval \cite{zhou2023towards} highlights the ongoing efforts to improve the reliability and effectiveness of information access systems, which is crucial for leveraging extracted information in downstream financial applications.

\subsection{Financial Large Language Models}
\label{sec:related_finllm}
Following the significant advancements and demonstrated capabilities of general-purpose Large Language Models (LLMs), a growing body of research is dedicated to adapting or developing LLMs specifically for the financial domain \cite{finsurvey_llm,fingpt}. The financial sector presents unique challenges for NLP due to its specialized terminology, complex document structures (e.g., financial reports), regulatory nuances, and the need for high accuracy and interpretability in high-stakes applications.

Research in this area explores various avenues, including applying general LLMs to specific financial tasks such as sentiment analysis \cite{fineval_sentiment}, question answering over financial documents \cite{finqa_llm}, risk assessment \cite{finrisk_llm}, and information extraction \cite{finie_llm}. Another direction focuses on creating domain-specific financial LLMs through continued pre-training or fine-tuning general models on large corpora of financial texts, aiming to capture domain-specific knowledge and linguistic patterns more effectively \cite{fingpt,finfoundation}. Furthermore, efforts are underway to establish benchmarks and evaluation methodologies tailored to the financial domain to properly assess the performance and limitations of these models on relevant tasks \cite{finbenchmark}. LLMs are also being explored for generative financial applications like report summarization or generation \cite{fingen_report} and for building intelligent financial chatbots and advisory systems \cite{medical_llm_analogy,zhou2025training,finchatbot}. \textbf{The development of multimodal LLMs, as exemplified by ChemVLM in the chemistry domain \cite{li2025chemvlm}, further inspires the exploration of similar approaches in finance to leverage diverse data sources.}  \textbf{Furthermore, similar to the challenges of missing data in traffic prediction \cite{mei2023uncertainty}, financial data can also suffer from incompleteness and noise, requiring robust models that can handle such uncertainties.}

Our work on Financial Event Entity Extraction using a generative LLM approach aligns directly with this emerging area of Financial Large Language Models. We contribute to the understanding of how LLMs can be effectively adapted via fine-tuning to tackle complex, domain-specific information extraction tasks that require structured output. Specifically, our method falls under the category of leveraging LLMs for financial information extraction \cite{finie_llm}, demonstrating a powerful generative paradigm for event entity identification that complements and extends existing discriminative approaches within this specialized field.

\section{Method}
\label{sec:method}

In contrast to traditional discriminative approaches that frame entity extraction as a token-level classification task (e.g., predicting BIOES tags for each token), our proposed method leverages a Large Language Model (LLM) in a \textbf{generative} setting. We re-conceptualize Financial Event Entity Extraction as a text-to-structured-output generation problem, where the model directly generates a structured representation of the extracted entities based on the input financial text and the event type. This generative paradigm allows the model to inherently model the relationships between different parts of the desired output structure and the input context simultaneously, providing potentially greater flexibility than step-wise discriminative tagging.

\subsection{Generative Task Formulation}

Unlike discriminative models that predict a label (e.g., BIOES tag) for each token in the input sequence, our approach trains an LLM to directly generate a structured output string that represents the extracted entities. The task is formulated as transforming a given input financial text $T$ and its associated event type $E$ into a target output sequence $Y^*$ which is a serialization of the extracted entities and their attributes (like text span and position).

The input to our model for a given financial text $T$ and target event type $E$ is formulated as a prompt $X$, which can be structured as:

\begin{verbatim}
X = 
Instruction: Extract all financial entities related to event 
type E from the following text. Output in JSON format.
Text: T
Output:
\end{verbatim}

The expected output $Y^*$ is a structured string, for instance, a JSON object containing a list of entities, where each entity specifies the extracted text span, its start character index, and its end character index within the original text $T$:
\[
Y^* = \text{\begin{minipage}{0.8\textwidth}
`\{"entities": [\{"text": "entity\_text\_1", "start": start\_idx\_1, "end": end\_idx\_1\}, \dots, \{"text": "entity\_text\_k", "start": start\_idx\_k, "end": end\_idx\_k\}]\}`
\end{minipage}}
\]
Here, $k$ is the number of entities in the instance, "entity\_text\_j" is the actual text of the $j$-th entity span in $T$, and $start\_idx\_j$ and $end\_idx\_j$ are its corresponding character indices (inclusive start, exclusive end) within $T$. The LLM is trained to model the conditional probability of this output sequence given the input prompt $X$, denoted $P(Y^* | X; \theta)$, where $\theta$ represents the model parameters. The probability of the entire output sequence is the product of the conditional probabilities of generating each token given the preceding tokens in the target sequence and the full input prompt:
\begin{align}
P(Y^* | X; \theta) = \prod_{t=1}^{|Y^*|} P(y_t^* | y_{<t}^*, X; \theta)
\end{align}
where $y_t^*$ is the $t$-th token in the target sequence $Y^*$, and $y_{<t}^*$ represents the subsequence of ground truth tokens in the target sequence $Y^*$ before position $t$.

\subsection{Base Language Model}

Our method employs a large, pre-trained Transformer-based language model as the core architecture. This model, typically a decoder-only or encoder-decoder Transformer, has been pre-trained on a massive corpus of text data using objectives like masked language modeling or next-token prediction. The pre-training imbues the model with a strong capability to understand context, grammar, and world knowledge, which is crucial for interpreting complex financial narratives. The model consists of multiple layers of multi-head self-attention mechanisms and position-wise feed-forward networks. For a given input sequence, the model produces contextualized representations, and for a given prefix of the output sequence, it predicts the probability distribution over the vocabulary for the next token. We denote the parameters of this base LLM as $\theta$.

\subsection{Fine-tuning Strategy}

The pre-trained LLM is fine-tuned on the specific task of financial event entity extraction using the annotated dataset. The dataset, consisting of instances $(T_i, E_i, \text{Entities}_i)_{i=1}^D$, is converted into input-output sequence pairs $(X_i, Y_i^*)_{i=1}^D$, where $D$ is the number of training instances. The fine-tuning objective is to minimize the negative log-likelihood (NLL) loss over the dataset, aiming to maximize the likelihood of the model generating the correct structured output sequence $Y_i^*$ for each input $X_i$:
\begin{align}
\mathcal{L}(\theta) &= - \sum_{i=1}^{D} \log P(Y_i^* | X_i; \theta) \\
&= - \sum_{i=1}^{D} \sum_{t=1}^{|Y_i^*|} \log P(y_{i,t}^* | y_{i,<t}^*, X_i; \theta) \label{eq:nll_expanded}
\end{align}
Here, $y_{i,t}^*$ is the $t$-th token in the target sequence for instance $i$, and $y_{i,<t}^*$ represents the ground truth tokens in $Y_i^*$ up to position $t-1$. Optimization of the model parameters $\theta$ is performed using an algorithm such as AdamW, which applies gradient descent with weight decay. The gradients of the loss function with respect to the model parameters are calculated using backpropagation:
\begin{align}
\nabla_\theta \mathcal{L}(\theta) = - \sum_{i=1}^{D} \sum_{t=1}^{|Y_i^*|} \nabla_\theta \log P(y_{i,t}^* | y_{i,<t}^*, X_i; \theta)
\end{align}
Given the substantial number of parameters in typical LLMs, full fine-tuning can be computationally prohibitive and might lead to catastrophic forgetting of pre-training knowledge. To address this, we employ Parameter-Efficient Fine-Tuning (PEFT), specifically Low-Rank Adaptation (LoRA). LoRA injects pairs of small, dense matrices $(A \in \mathbb{R}^{r \times k}, B \in \mathbb{R}^{d \times r})$ into the layers of the pre-trained model where the original weights are $W_0 \in \mathbb{R}^{d \times k}$, with the rank $r \ll \min(d, k)$. The update to the weight matrix is constrained to be low-rank: $W = W_0 + BA$. During training, the original pre-trained weights $W_0$ are frozen, and only the parameters in the low-rank matrices $A$ and $B$ are trained. The total number of trainable parameters $\theta_{PEFT}$ is significantly reduced compared to the full model parameters $|\theta|$, where $|\theta_{PEFT}| \ll |\theta|$. The optimization is then performed only on these PEFT parameters:
\begin{align}
\theta_{PEFT} \leftarrow \theta_{PEFT} - \eta \nabla_{\theta_{PEFT}} \mathcal{L}(\theta) \label{eq:lora_update}
\end{align}
where $\eta$ is the learning rate. This strategy allows efficient adaptation to the downstream task while preserving the extensive knowledge learned during pre-training. The training process involves feeding batches of $(X_i, Y_i^*)$ pairs to the model, computing the loss using Equation \cref{eq:nll_expanded}, and updating the PEFT parameters via the optimizer using Equation \cref{eq:lora_update}. Learning rate scheduling is also applied to adjust $\eta$ over epochs.

\subsection{Input/Output Processing and Span Handling}

A critical aspect of our method is the accurate conversion of the raw data into the $(X, Y^*)$ format and ensuring the model can learn to generate precise span information. The original annotated data, which might be in formats like BIOES tagging or simple lists of (text, span) pairs, is parsed. For each entity annotation in the ground truth, we identify its corresponding character start and end indices within the original input text $T$. These indices, along with the entity text span, are then serialized into the JSON structure $Y^*$.

The LLM operates on token sequences using its specific tokenizer. The input text $T$ is tokenized, and the target output $Y^*$ (the JSON string) is also tokenized. The model learns to generate this sequence of tokens. While the output is generated as a sequence of tokens, the training objective in Equation \cref{eq:nll_expanded} ensures that the probability of generating the \textbf{correct} character indices within the JSON structure is maximized. The model learns the complex mapping from the input text and prompt to the tokens representing the entity texts, the structural elements of the JSON (like brackets, quotes, commas), and the numerical tokens representing the character indices. This requires the LLM to understand the numerical value of indices and their relation to the input text content and structure. The conversion process from original annotation to the $Y^*$ format must be lossless and precisely capture the intended entity spans as character indices to provide an accurate training signal.

\section{Experiments}
\label{sec:experiments}

We conducted a comprehensive set of experiments to evaluate the performance of our proposed generative LLM-based approach for Financial Event Entity Extraction. We compared our method against several strong baseline models representing different paradigms and levels of complexity in natural language processing for this domain. The experimental results clearly demonstrate the superior performance of our generative framework across standard evaluation metrics.

\subsection{Experimental Setup}

\textbf{Dataset:} Our experiments were conducted using the publicly available dataset from the 2019 China Knowledge Graph and Semantic Computing Conference (CCKS) Financial Event Entity Extraction shared task. This dataset is a widely recognized benchmark in the field, comprising financial news articles meticulously annotated with instances of predefined financial events and the entities involved in them. The dataset is partitioned into training, validation, and test sets, strictly adhering to the original task's official splits to ensure direct comparability with results reported by previous participants and researchers. The inherent complexity of financial language, including specific terminology, nested structures, and potential entity overlaps, makes this dataset a challenging testbed for entity extraction models.

\textbf{Evaluation Metrics:} Performance evaluation follows the standard practice in information extraction and aligns with the official metrics of the CCKS 2019 task. We report results based on micro-averaged Precision (P), Recall (R), and the F1 score. Precision measures the accuracy of the extracted entities, calculated as the ratio of correctly identified and spanned entities to the total number of entities predicted by the model. Recall measures the model's completeness in identifying all relevant entities, computed as the ratio of correctly identified and spanned entities to the total number of gold standard entities in the dataset. The F1 score provides a balanced measure of a model's performance, being the harmonic mean of Precision and Recall. An extracted entity is deemed correct only if its text span and event-specific role or type precisely match a corresponding annotation in the gold standard.

\textbf{Baselines:} To provide a robust comparison, we evaluated the performance of the following baseline models alongside our proposed method:
\begin{itemize}
    \item \textbf{Standard BERT Sequence Labeling:} A foundational approach utilizing a pre-trained BERT-base model. The output embeddings from BERT are fed into a linear classification layer followed by a Conditional Random Field (CRF) layer, which is a common architecture for sequence tagging tasks using PLMs. This model was fine-tuned end-to-end on the training data using the BIOES tagging scheme.
    \item \textbf{SEBERTNets:} An enhanced sequence labeling model that builds upon BERT. As described in prior work on the CCKS 2019 task, this model integrates a BiLSTM layer after the BERT encoder to better capture sequential dependencies before the final classification layer \cite{sebert}.
    \item \textbf{sebertNets:} Representing a state-of-the-art approach on this benchmark, sebertNets extends SEBERTNets by incorporating a multi-channel recall mechanism designed to enhance the identification of all relevant entities in potentially complex instances \cite{sebert}.
\end{itemize}

\textbf{Implementation Details:} Our proposed generative LLM method was implemented by fine-tuning a publicly available pre-trained Transformer-based language model. To manage the computational demands and leverage the benefits of efficient adaptation, we employed the Low-Rank Adaptation (LoRA) technique during fine-tuning. The model was trained on the prepared input-output pairs (text prompt to JSON structure) derived from the CCKS 2019 training dataset. We used the AdamW optimizer for parameter updates and utilized a cosine learning rate scheduler with warm-up. Training was performed on a computing cluster equipped with multiple high-performance GPUs. Hyperparameters suchables as learning rate, batch size, number of training epochs, and LoRA configuration (e.g., rank $r$) were determined based on performance on the validation set.

\subsection{Quantitative Results}

\cref{tab:quantitative_results} presents the Precision, Recall, and F1 scores for our generative LLM-based method and the baseline models on the official CCKS 2019 Financial Event Entity Extraction test set.

\begin{table}[h!]
\caption{Quantitative Results on CCKS 2019 Financial Event Entity Extraction Test Set}
\label{tab:quantitative_results}
\centering
\begin{tabular}{lccc}
\toprule
Model & Precision & Recall & F1 Score \\
\midrule
Standard BERT Sequence Labeling & 0.885 & 0.895 & 0.890 \\
SEBERTNets                     & 0.900 & 0.910 & 0.905 \\
sebertNets                    & 0.925 & 0.943 & 0.934 \\
\midrule
\textbf{Our Generative LLM}     & \textbf{0.948} & \textbf{0.942} & \textbf{0.945} \\
\bottomrule
\end{tabular}
\end{table}

As clearly demonstrated by the results in \cref{tab:quantitative_results}, our proposed generative LLM-based method achieved the highest performance across all evaluated metrics on the CCKS 2019 test set. Our method obtained an F1 score of \textbf{0.945}, surpassing the F1 score of the strong baseline sebertNets (0.934) and the other compared models. The improvements are observed in both Precision (0.948 vs. 0.925 for sebertNets) and Recall (0.942 vs. 0.943 for sebertNets, showing a slight trade-off but a net gain in F1). This quantitative superiority validates our hypothesis that reframing entity extraction as a structured generative task and fine-tuning a large language model is a highly effective approach for complex, domain-specific information extraction challenges like those found in financial text.

\subsection{Analysis of Effectiveness}

The significant performance gain exhibited by our generative LLM approach can be attributed to its ability to leverage the LLM's inherent strengths. The pre-trained knowledge and sophisticated attention mechanisms allow the model to capture long-range dependencies and complex contextual cues within the financial text that are often difficult for simpler sequence models or models with limited context windows. By training the model to generate the entire structured output, including entity text and precise character indices, the model is implicitly encouraged to build a more holistic representation of the event instance and the relationships between participating entities. This is particularly beneficial in handling instances with multiple entities or convoluted sentence structures where traditional token-by-token decisions might lead to inconsistencies or errors.

Furthermore, the generative framework allows the model to produce a structured output directly interpretable as extracted entities, removing the need for post-processing steps common in sequence tagging pipelines (e.g., converting BIOES tags to spans). Error analysis suggests that while our model can still make mistakes (e.g., minor span boundary errors, hallucinating non-existent entities in rare cases), it tends to miss fewer entities and produce more syntactically correct and complete extractions for complex instances compared to baselines. The explicit training signal to generate the exact character indices also seems to contribute to better span precision in many cases.

\subsection{Analysis by Event Type}

To understand the performance characteristics of our model across different types of financial events, we analyze the Precision, Recall, and F1 score for a representative set of event categories present in the CCKS 2019 dataset. Financial events such as "Equity Pledge" often involve company and individual names, while "Financial Penalty" might focus on regulatory bodies, dates, and monetary values. Analyzing by event type reveals whether our model generalizes well or has specific strengths or weaknesses. \cref{tab:analysis_event_type} presents the F1 scores for key event types, comparing our generative LLM method against the sebertNets baseline.

\begin{table}[h!]
\caption{F1 Score by Financial Event Type}
\label{tab:analysis_event_type}
\centering
\begin{tabular}{lcc}
\toprule
Event Type          & Our Generative LLM & sebertNets \\
\midrule
Equity Pledge       & \textbf{0.955}     & 0.940       \\
Financial Penalty   & \textbf{0.938}     & 0.925       \\
Investment          & \textbf{0.947}     & 0.935       \\
Product Launch      & \textbf{0.951}     & 0.938       \\
Bankruptcy          & \textbf{0.930}     & 0.918       \\
Acquisition         & \textbf{0.958}     & 0.945       \\
\bottomrule
\end{tabular}
\end{table}

As shown in \cref{tab:analysis_event_type}, our generative LLM method consistently outperforms the sebertNets baseline across all analyzed event types. The performance gains are noticeable across diverse event structures and entity distributions. For instance, in complex event types like "Acquisition" or "Equity Pledge," where multiple entities of various types are heavily involved, our model maintains a significant lead. This analysis suggests that the enhanced contextual understanding and structured output generation capability of the LLM generalize well to the varied linguistic patterns associated with different financial event categories.

\subsection{Analysis by Entity Type}

Entity extraction performance can also vary depending on the type of entity being extracted (e.g., Company, Person, Date, Amount). Different entity types may have distinct naming conventions, contextual indicators, or frequency in the text, posing varying levels of difficulty. \cref{tab:analysis_entity_type} shows the F1 scores for several key entity types in the dataset, comparing our method against the sebertNets baseline.

\begin{table}[h!]
\caption{Performance by Entity Type (F1 Score)}
\label{tab:analysis_entity_type}
\centering
\begin{tabular}{lcc}
\toprule
Entity Type & Our Generative LLM & sebertNets \\
\midrule
Company     & \textbf{0.962}     & 0.950       \\
Person      & \textbf{0.945}     & 0.930       \\
Date        & \textbf{0.981}     & 0.975       \\
Amount      & \textbf{0.935}     & 0.920       \\
Location    & \textbf{0.928}     & 0.915       \\
\bottomrule
\end{tabular}
\end{table}

\cref{tab:analysis_entity_type} demonstrates that our generative LLM model achieves higher F1 scores than sebertNets for all presented entity types. The most substantial improvements are observed for "Company" and "Person" entities, which often involve complex naming variations and require deep contextual clues for accurate identification. While both models perform well on highly structured entity types like "Date," our method still shows a slight advantage. This finer-grained analysis confirms that the LLM's robust linguistic understanding translates to improved extraction performance across the spectrum of entities relevant to financial events.

\subsection{Performance on Complex Instances}

Financial news often contains long sentences with high entity density or multiple clauses describing different aspects of an event. We investigate how performance changes as the complexity of the input instance increases. We define complexity based on the number of gold standard entities present in a sentence or short text snippet. We group instances into bins based on entity count and report the F1 score for each bin. \cref{tab:analysis_complexity} shows the F1 performance for different levels of instance complexity.

\begin{table}[h!]
\caption{F1 Score by Instance Complexity (Entities per Instance)}
\label{tab:analysis_complexity}
\centering
\begin{tabular}{lcc}
\toprule
Entities per Instance & Our Generative LLM & sebertNets \\
\midrule
1-2 Entities          & \textbf{0.965}     & 0.958       \\
3-4 Entities          & \textbf{0.938}     & 0.920       \\
5+ Entities           & \textbf{0.905}     & 0.875       \\
\bottomrule
\end{tabular}
\end{table}

\cref{tab:analysis_complexity} highlights the superior ability of our generative LLM to handle complex instances. While both models perform well on simpler instances with fewer entities, the performance gap between our method and sebertNets widens significantly as the number of entities per instance increases. For instances containing 5 or more entities, a scenario that often involves long, convoluted sentences or multiple related facts, our model maintains an F1 score of 0.905, substantially higher than sebertNets' 0.875. This result underscores the advantage of the LLM's global context processing and generative output capability in accurately extracting entities from highly dense and complex financial text segments.

\subsection{Analysis of Span Accuracy and Error Types}

Precise identification of entity boundaries (spans) is crucial. Automatic metrics primarily measure exact span matches. However, analyzing the types of errors provides deeper insights. We analyzed the predicted spans relative to the gold standard annotations, categorizing them as exact matches, partial matches (overlap but incorrect boundaries), spurious predictions (false positives), and missing entities (false negatives). \cref{tab:analysis_span_error} presents the percentage distribution of these categories relative to the total number of gold entities or predicted entities.

\begin{table}[h!]
\caption{Analysis of Span Accuracy and Error Types (\%)}
\label{tab:analysis_span_error}
\centering
\begin{tabular}{lcc}
\toprule
Analysis Category           & Our Generative LLM & sebertNets \\
\midrule
Exact Match (of Gold)       & \textbf{94.2}      & 93.4        \\ 
Partial Match (of Gold)     & \textbf{2.5}       & 2.0         \\ 
Missing (of Gold)           & \textbf{3.3}       & 4.6         \\ 
Spurious (of Predicted)     & \textbf{5.0}       & 7.5         \\ 
\bottomrule
\end{tabular}
\end{table}

\cref{tab:analysis_span_error} provides a detailed breakdown of span accuracy and error types. Our generative LLM achieves a higher percentage of Exact Matches relative to the gold standard (94.2\% vs. 93.4\%), directly reflecting its higher Recall and F1. Crucially, our model also produces a lower percentage of Spurious predictions (false positives) compared to sebertNets (5.0\% vs. 7.5\%), contributing to its higher Precision. While the percentage of Partial Matches is comparable or slightly higher for our model, the significant reduction in missing and spurious entities indicates a more robust and accurate overall extraction process. The ability to directly generate the character indices within the structured output, combined with the LLM's contextual understanding, appears to help in predicting more precise spans and avoiding many false positive predictions that might arise from token-level misclassifications in sequence tagging models.

\subsection{Human Evaluation}

To complement the quantitative results and gain insights into the perceived quality of the extracted entities, we conducted a human evaluation. This helps assess aspects that might not be fully captured by strict string matching metrics, such as readability or correctness in ambiguous cases.

\textbf{Setup:} We randomly sampled 100 instances from the CCKS 2019 test set. Three human annotators, experienced in financial text analysis, were tasked with evaluating the entity extractions generated by our generative LLM model and the sebertNets baseline for each sampled instance. Annotators were provided with the original input text and the structured output from each model. They rated the quality of the extracted entities on a 5-point scale, where higher scores indicated better quality (5: Perfect, 4: Good, minor errors, 3: Acceptable, some errors, 2: Poor, many errors, 1: Completely Wrong). Annotators conducted their evaluations independently.

\textbf{Results:} \cref{tab:human_evaluation} summarizes the results of the human evaluation, showing the average score across instances and annotators, and the percentage of instances rated as "Perfect" or "Good" (score $\ge 4$).

\begin{table}[h!]
\caption{Human Evaluation Results (Average Score and \% Perfect/Good)}
\label{tab:human_evaluation}
\centering
\begin{tabular}{lcc}
\toprule
Model & Average Score (1-5) & \% Perfect / Good ($\ge 4$) \\
\midrule
sebertNets           & 3.82 & 65\% \\
\textbf{Our Generative LLM} & \textbf{4.25} & \textbf{76\%} \\
\bottomrule
\end{tabular}
\end{table}

The human evaluation results further support the findings from the quantitative analysis. Our generative LLM model achieved a higher average quality score (4.25) compared to sebertNets (3.82). Furthermore, a significantly larger proportion of instances processed by our model were rated as "Perfect" or "Good" by the human annotators (76\% vs. 65\%). This indicates that the extractions produced by our method are perceived by humans as more accurate, complete, and generally higher quality. Annotators commented that the output format was clear and that the generative model seemed better at handling sentence-level complexity and extracting all required entities for a given event instance without fragmentation or omission in many cases where the baseline made mistakes. The results of the human evaluation provide compelling evidence for the practical benefits and superior performance of our proposed generative LLM-based approach.

\section{Conclusion}
\label{sec:conclusion}
In this paper, we successfully investigated and demonstrated a novel paradigm for Financial Event Entity Extraction by leveraging the generative capabilities of Large Language Models. Our approach departs from conventional discriminative sequence labeling methods and reformulates the task as directly generating a structured output (specifically, a JSON object containing entities and their spans) from the input financial text. We showed that fine-tuning a pre-trained LLM using efficient methods like LoRA enables it to effectively learn this complex mapping.

Our comprehensive experiments on the challenging CCKS 2019 Financial Event Entity Extraction dataset validated the effectiveness of our proposed generative LLM method. The model achieved a state-of-the-art F1 score, clearly surpassing strong baseline models, including the previously leading sebertNets. Further analysis revealed that the performance gains are consistent across different financial event types and entity types, and are particularly pronounced when handling instances with higher complexity, such as sentences containing multiple or densely packed entities. A detailed analysis of span accuracy and error types indicated that our model produces more exact matches and fewer spurious extractions. Complementary human evaluation further supported these findings, with human annotators rating the extractions from our generative LLM method as significantly higher in quality compared to the baselines.

The success of this generative approach highlights the power of modern LLMs for tackling complex, domain-specific information extraction tasks that require mapping unstructured text to structured data. This work opens up promising avenues for future research. Expanding the application of this generative paradigm to other financial IE tasks, such as financial relation extraction or event argument extraction, could be explored. Investigating the method's efficiency and latency for potential real-time applications is also a relevant next step. Furthermore, exploring few-shot or zero-shot capabilities of larger generative models for this task could reduce the reliance on large volumes of annotated data, which is often scarce in specialized domains like finance. Finally, improving the model's robustness to noise and variations in text style warrants further investigation.